%% file: iclr2027_conference.tex
\documentclass{article} 
\usepackage{iclr2027_conference,times}

\input{math_commands.tex}

\usepackage{hyperref}
\usepackage{url}
\usepackage{booktabs}       
\usepackage{amsfonts}       
\usepackage{nicefrac}       
\usepackage{microtype}      
\usepackage{xcolor}         
\usepackage{amsmath}
\usepackage{subcaption}
\usepackage{graphicx}
\usepackage{multirow}
\usepackage{placeins}
\usepackage{capt-of}
\usepackage{tabularx}
\usepackage[table]{xcolor}
\usepackage{pifont}
\usepackage{wrapfig}
\usepackage{needspace}
\usepackage{adjustbox}
\usepackage{marvosym}
\usepackage{fontawesome5}
\usepackage{algorithm}
\usepackage{algpseudocode}

\newsavebox{\ablationleftbox}
\newsavebox{\ablationcaptionbox}

\newlength{\ablationcolheight}
\newlength{\ablationimageheight}
\newlength{\ablationleftwidth}
\newlength{\ablationrightwidth}

\definecolor{lightgray}{gray}{0.9}
\definecolor{linecolor}{rgb}{0.82, 0.94, 0.75}
\definecolor{mamba}{RGB}{153, 151, 239}
\definecolor{kaiming-green}{RGB}{57,181,74} 
\definecolor{pretty-blue}{RGB}{0, 113, 188}

\hypersetup{
    colorlinks=true,
    linkcolor=red,
    filecolor=magenta,      
    urlcolor=magenta,
    citecolor=kaiming-green
}

\newcommand{\ourmethod}{TurboVLA }

\newcommand{\cmark}{\textcolor{green!70!black}{\ding{51}}}
\newcommand{\xmark}{\textcolor{red}{\ding{55}}}

\title{TurboVLA: Real-Time Vision-Language-Action Model at 32 Hz on an RTX 4090 with \textless 1 GB VRAM}

\author{
    Hengyi Xie\thanks{Equal contribution, listed alphabetically by surname.~\textsuperscript~\textsuperscript{$\dagger$}Project Lead.}~, Chenfei Yao$^{*}$, Xianjin Wu,
    \textbf{Yingying Zhu, Dingkang Liang\textsuperscript{$\dagger$},} \\
    ~\textbf{Xiang Bai, Han Ding}\\\\
    ~Huazhong University of Science and Technology\\
    ~\texttt{\{hengyi\_xie, yaochenfei, dkliang, xbai\}@hust.edu.cn}
    \\
    \begin{tabular}{@{}c@{\hspace{0.55em}}l@{}}
    {\color{black}\faGithub}
    &
    \url{https://github.com/H-EmbodVis/TurboVLA}
    \\[2pt]
    \raisebox{-0.2em}{
    \includegraphics[height=1.1em]{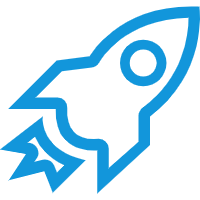}
    }
    &
    \url{https://H-EmbodVis.github.io/TurboVLA}
    \end{tabular}
}

\iclrfinalcopy 
\begin{document}

\maketitle
\fancyhead[L]{}

\vspace{-20pt}
\begin{minipage}{\textwidth}
    \centering
    \includegraphics[
        width=0.88\textwidth
    ]{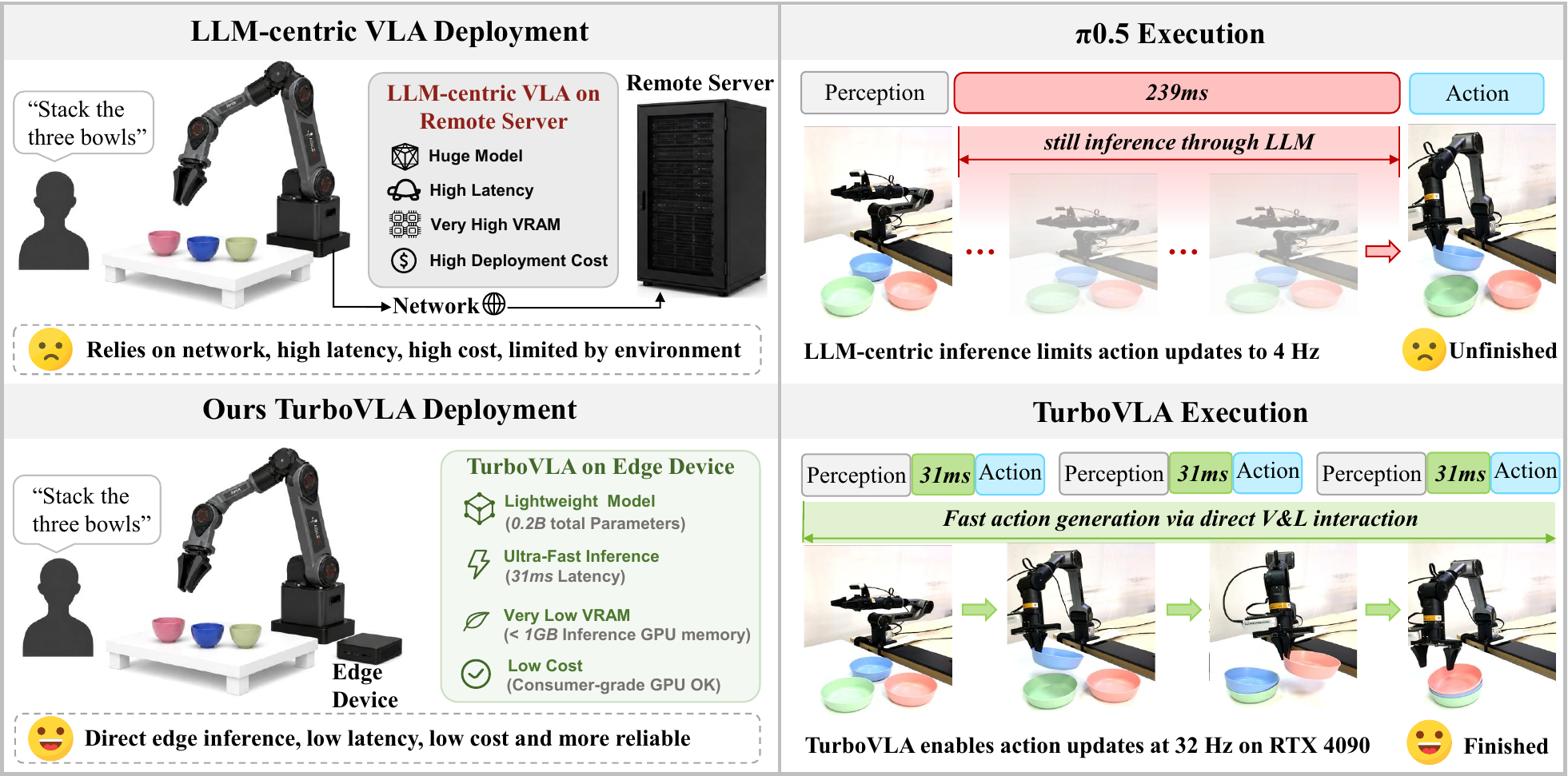}

    \captionsetup{
        font=small,
        labelfont=bf,
        justification=centering,
        singlelinecheck=false,
        skip=4pt,
        width=0.90\textwidth
    }
    \captionof{figure}{
    TurboVLA enables compact local deployment and real-time language-conditioned manipulation with 0.2B parameters, 0.9\,GiB inference VRAM and 31.2\,ms policy latency.
    }
    \label{fig:promo}
\end{minipage}

\par\vspace{1.2em}

\begin{abstract}
Vision-language-action (VLA) models commonly adopt an LLM-centric
\mbox{$V\rightarrow \boldsymbol{L}\rightarrow A$} pathway, processing visual observations and language instructions through a large language model before predicting robot actions. Although effective, this design incurs substantial computation and memory overhead. In this work, we introduce \textbf{TurboVLA}, a compact VLA architecture built on a direct \mbox{$\boldsymbol{V}+\boldsymbol{L}\rightarrow A$} mapping. Instead of using a large language model as the central interface between perception and action, \ourmethod independently encodes visual observations and language instructions, directly exchanges information between them through lightweight bidirectional vision-language interaction, and predicts continuous action chunks with a compact decoder. This simple design directly constructs task-conditioned representations while avoiding the overhead of an LLM-centric execution pathway. On LIBERO, \ourmethod achieves 97.6\% average success with only 0.2B parameters, 31.2\,ms inference latency, and 0.9\,GiB inference VRAM on a consumer-grade RTX 4090. Notably, a 0.4\,B \ourmethod achieves 88.06\% success on RoboTwin 2.0, even matching or outperforming substantially larger VLA policies. These results demonstrate that the simple \mbox{$\boldsymbol{V}+\boldsymbol{L}\rightarrow A$} design of TurboVLA can achieve high performance without requiring an LLM-centric execution pathway, offering a new perspective on how vision, language, and action can be connected for efficient robotic manipulation.
\end{abstract}

\section{Introduction}

Vision-language-action (VLA) models have become a powerful framework for language-conditioned robotic manipulation, connecting visual observations, language instructions, and robot actions within a unified policy~\citep{brohan2022rt, team2024octo, black2024pi_0, pmlr-v305-black25a, liu2025rdt, li2024cogact,fu2025orion,fang2026towards}. A common design is to place a large language model at the center of this process. Such systems effectively follow an indirect \mbox{$V\rightarrow \boldsymbol{L} \rightarrow A$} pathway: visual observations and language instructions are mapped into LLM-aligned tokens, which are subsequently decoded into robot actions~\citep{driess2023palm, zitkovich2023rt, kim2024openvla}. This design transfers broad semantic knowledge from large-scale pretraining to robot control and supports semantic generalization and high-level reasoning.

\begin{figure*}[t]
    \centering
    \includegraphics[width=0.98\textwidth]{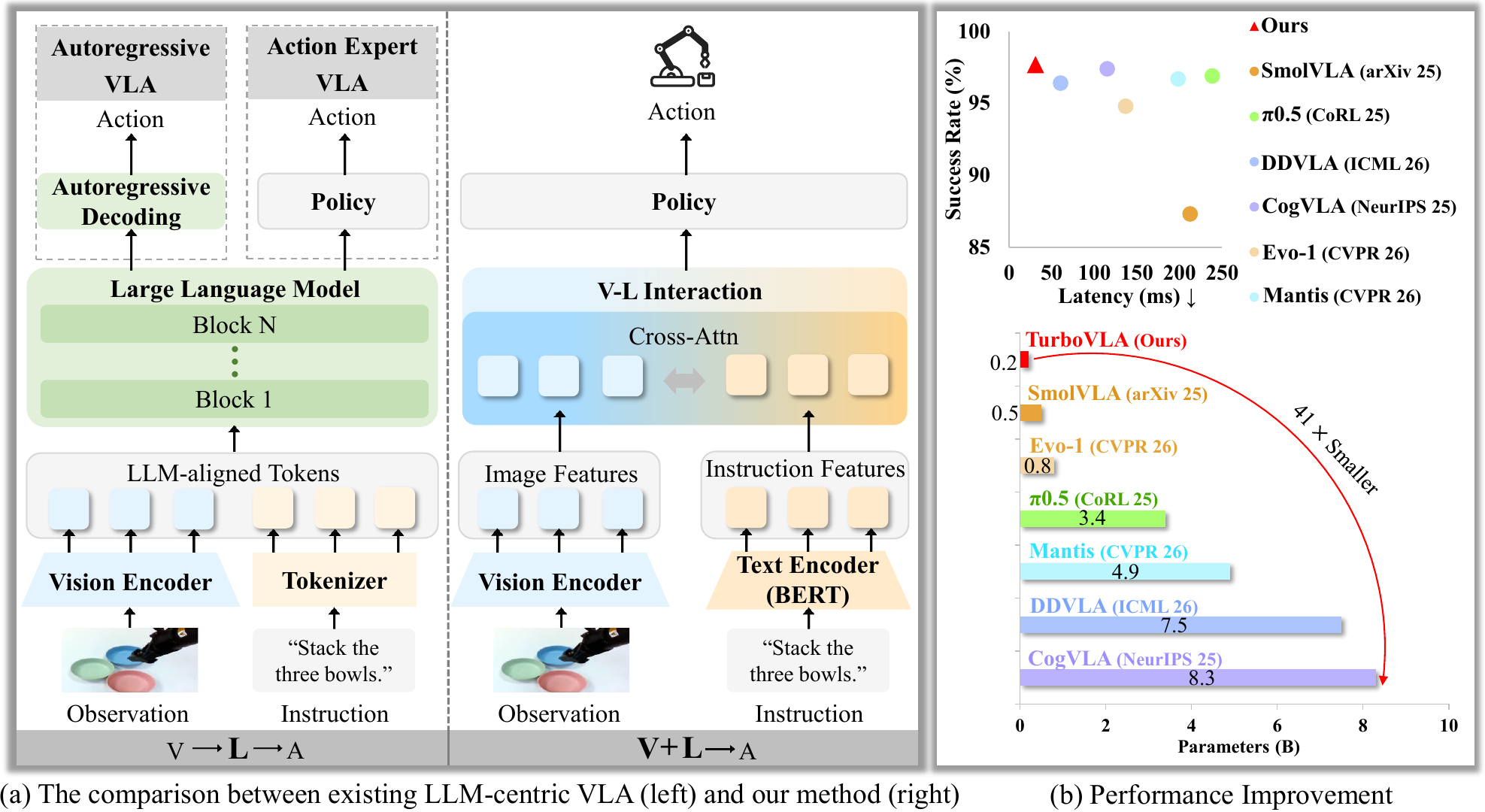}
    \caption{
    (a) LLM-centric VLA predicts actions from large-language-model representations, whereas \ourmethod directly fuses visual and instruction features for continuous control. (b) \ourmethod achieves highly competitive LIBERO performance with substantially lower latency and model scale.
    }
    \label{fig:teaser}
\end{figure*}

LLM-centric VLA models, however, can impose a substantial computational bottleneck on real-time robotic execution, where low latency is essential for responsive interaction, high-throughput manipulation, and deployment on resource-constrained robotic platforms. As summarized in Fig.~\ref{fig:teaser}(a), these models typically generate actions either autoregressively as action tokens~\citep{kim2024openvla,zitkovich2023rt} or through action experts that predict continuous action chunks~\citep{black2024pi_0, pmlr-v305-black25a, kim2025fine, li2024cogact, shukor2025smolvla}. Although action-expert designs avoid the token-by-token decoding overhead of autoregressive language generation, they still rely on a large language or multimodal backbone to process visual observations and instructions. The resulting computational and memory overhead increases inference latency and limits control frequency. This raises a more fundamental question: \emph{how to design a simple, elegant and efficient VLA that directly maps vision and language to actions for execution-level manipulation, without centering on a large language model?}

Our key observation is that execution-level control need not be centered on a large language model. When an instruction specifies the intended manipulation skill, the execution policy does not need to perform explicit high-level reasoning or subtask generation. Instead, the policy can primarily focus on grounding the instruction in the current visual observations and using this information to guide action. In LLM-centric VLA models, this interaction is mediated through a general-purpose language-model representation. We argue that a lightweight text encoder, such as BERT~\citep{devlin2019bert}, can capture execution-relevant instruction semantics, while compact cross-modal interaction allows visual and linguistic features to inform each other~\citep{lynch2020language, shridhar2022cliport, jang2022bc, mees2022calvin}. This motivates a direct vision-language interaction design that constructs control-oriented representations for continuous action prediction.

Therefore, we propose \textbf{TurboVLA}, a simple yet efficient \mbox{$\boldsymbol{V}+\boldsymbol{L}\rightarrow A$} model for real-time language-conditioned manipulation. As shown in Fig.~\ref{fig:teaser}(a), \ourmethod separately encodes visual observations and task instructions using a visual encoder and a lightweight text encoder. Inspired by the efficient cross-modal interaction in Grounding DINO~\citep{liu2024grounding}, \ourmethod uses a compact cross-attention module to fuse visual and linguistic features and then decodes the resulting representation into continuous action chunks~\citep{zhao2023learning}. By directly connecting visual and linguistic features to action prediction, \ourmethod avoids the overhead of an LLM-centric execution pathway, significantly reducing inference latency and GPU memory usage.

Extensive experiments show that \ourmethod offers a favorable performance--efficiency trade-off. On LIBERO~\citep{liu2023libero}, \ourmethod achieves 97.6\% average success, with 0.2B parameters, 31.2\,ms policy inference latency and \textless 1\,GB of inference VRAM on a consumer-grade RTX 4090. On RoboTwin 2.0~\citep{chen2025robotwin}, a 0.4B \ourmethod variant achieves 88.06\% average success, even matching or outperforming substantially larger LLM-centric VLA systems. Real-world experiments further validate the effectiveness and efficiency of TurboVLA. TurboVLA substantially reduces the computational and memory requirements for deploying language-conditioned manipulation policies in latency-sensitive and resource-constrained robotic systems. These results suggest that lightweight language encoding and direct vision-language interaction enable competitive and efficient execution-level manipulation without an LLM-centric policy. Our contributions are summarized as follows:
\vspace{-10pt}
\begin{itemize}
\item We revisit the computational and memory costs of LLM-centric VLA models and highlight how lightweight instruction encoding and direct bidirectional vision-language interaction can retain task-relevant language semantics for execution-level control, without placing a large language model between perception and action.
\item We propose TurboVLA, a simple and efficient \mbox{$\boldsymbol{V}+\boldsymbol{L}\rightarrow A$} model that separately encodes visual observations and task instructions, and allows the two modalities to jointly construct task-conditioned, control-oriented representations through direct bidirectional vision-language interaction for continuous action prediction.
\item We validate TurboVLA on simulation benchmarks and real-world manipulation tasks, demonstrating competitive manipulation performance with reduced computational and memory requirements. These results highlight the potential for more accessible robotic deployment and provide insight into efficient policy design beyond LLM-centric architectures.
\end{itemize}

\section{Related Work}

\textbf{Vision-language-action models.}
Vision-language-action (VLA) models map visual observations and task instructions to robot actions, often leveraging large-scale vision-language pretraining for semantic generalization. RT-1~\citep{brohan2022rt} demonstrates scalable transformer-based control, while RT-2~\citep{zitkovich2023rt} and OpenVLA~\citep{kim2024openvla} adapt pretrained vision-language models through action tokens. Continuous-control models such as $\pi_0$~\citep{black2024pi_0} and $\pi_{0.5}$~\citep{pmlr-v305-black25a} attach dedicated action experts to pretrained multimodal backbones. Other generalist policies explore reusable policy representations~\citep{team2024octo}, diffusion-based action generation~\citep{liu2025rdt,li2024cogact}, and foundation models for diverse embodiments~\citep{bjorck2025gr00t,bu2025univla,wang2025unified}. Recent advances further enrich policy representations through visual foresight~\citep{yang2026mantis}, predictive world knowledge~\citep{zhang2026dreamvla}, and latent reasoning~\citep{bai2026latent}. \ourmethod focuses on the architecture used for execution-level control: in contrast to LLM-centric policies, it separately encodes task text and visual observations, directly fuses their features, and predicts actions conditioned on robot state without a large generative language backbone.

\textbf{Efficient execution in VLA policies.}
Existing approaches improve VLA efficiency at several levels. Action-side methods improve action generation through parallel decoding or action-chunk prediction~\citep{kim2025fine,liang2025discrete,liu2025rdt}, as well as compact action tokenization~\citep{pertsch2025fast,liu2026oat}. Compact architectures such as TinyVLA~\citep{wen2025tinyvla}, SmolVLA~\citep{shukor2025smolvla}, and Evo-1~\citep{lin2026evo} reduce model scale or inference cost. A complementary line of work reduces backbone inference cost through quantization~\citep{xu2026qvla}, token reuse and pruning~\citep{xu2026vlacache,jiang2025better}, adaptive depth and structural pruning~\citep{yang2026efficientvla,wang2025specprune}, and distillation~\citep{jeon2026shallow}. Other approaches improve inference or execution efficiency through asynchronous action-chunk execution~\citep{black2026real}, streaming inference with horizon-aware flow sampling~\citep{lu2026faster}, and speculative inference~\citep{niu2026realtime}. These methods optimize computation or execution within existing policy frameworks. \ourmethod instead targets the representation pathway, replacing the large generative language backbone with compact encoders and direct vision-language interaction for control.

\textbf{Language interfaces for robot control.}
Natural-language instructions have long been used as task specifications for robotic manipulation. Early imitation-learning methods explored language grounding for instruction-conditioned robotic control~\citep{lynch2020language,stepputtis2020language}. CLIPort~\citep{shridhar2022cliport} and BC-Z~\citep{jang2022bc} further explored language-conditioned manipulation, while CALVIN~\citep{mees2022calvin} provides a benchmark for long-horizon language-conditioned control and HULC~\citep{mees2022matters} develops policies for this setting. \ourmethod investigates whether compact instruction encoding and direct vision-language interaction can match the performance of an LLM-centric representational pathway while retaining efficiency in the execution-level VLA setting.

\section{TurboVLA}
\begin{figure*}[t]
\centering
\includegraphics[width=0.95\textwidth]{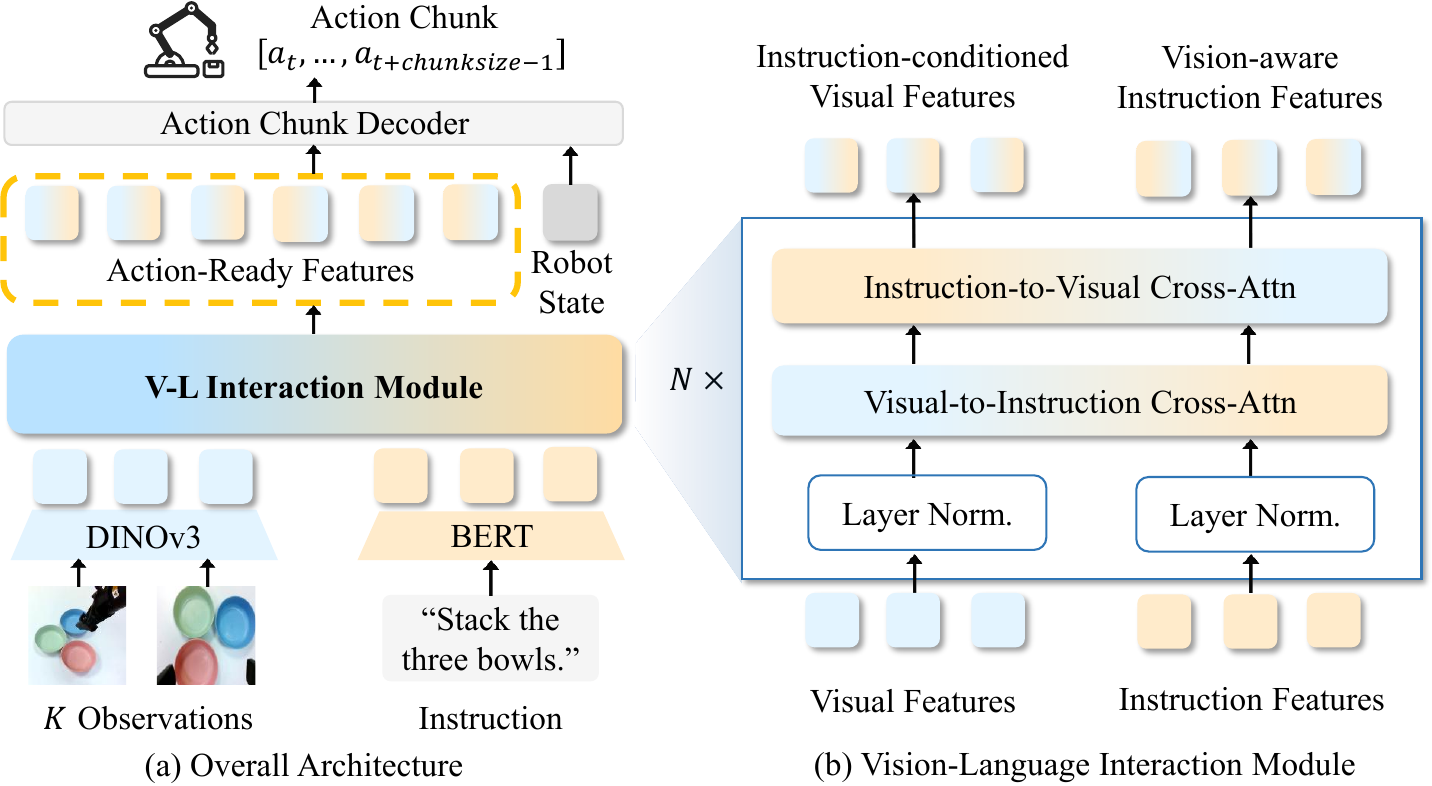}
\caption{
Overview of TurboVLA. (a) TurboVLA simply yet efficiently maps visual observations and language instructions to continuous action chunks through compact modality encoders, vision-language interaction, and an action chunk decoder. (b) The interaction module is designed to be as simple as possible. It uses stacked bidirectional cross-attention to produce vision-aware instruction features and instruction-conditioned visual features.
}
\label{fig:method}
\end{figure*}

We introduce \textbf{TurboVLA}, a simple and elegant \mbox{$\boldsymbol{V}+\boldsymbol{L}\rightarrow A$} model for execution-level language-conditioned manipulation. As illustrated in Fig.~\ref{fig:method}(a), TurboVLA separately encodes observations and instructions, allows the two modalities to exchange information through bidirectional cross-modal interaction, and combines the resulting representation with the robot state to predict continuous action chunks. This design preserves task-relevant semantics through a lightweight language encoder without placing a large language model as the central interface between perception and action, enabling competitive manipulation performance with efficient inference.

\subsection{Problem Formulation and Approach Overview}
\label{sec:problem_formulation}

At execution step $t$, a language-conditioned manipulation policy receives multi-view visual observations $\mathcal{O}_t$, a task instruction $x$, and the current robot state $s_t$, and predicts a chunk of $H$ continuous actions $\hat{\mathbf{A}}_t=\pi_{\theta}(\mathcal{O}_t,x,s_t)\in\mathbb{R}^{H\times d_a}$, where $d_a$ is the action dimension. The predicted actions are executed sequentially before the policy observes the next state and predicts a new action chunk.

\textbf{LLM-centric VLA implementation.}
Existing LLM-centric VLA models process visual observations and task instructions through a pretrained large language model:
\begin{equation}
H_t^L
=
F_L\!\left(
\left[
P_v\!\left(E_v(\mathcal{O}_t)\right);
\operatorname{Tok}(x)
\right]
\right),
\end{equation}
where $E_v$ is the visual encoder, $P_v$ projects visual features into the language-model embedding space, $\operatorname{Tok}(x)$ embeds the instruction, and $F_L$ denotes the large language model. The resulting multimodal representation supports either autoregressive action-token prediction ~\citep{kim2024openvla} or continuous action generation through an action module~\citep{black2024pi_0,pmlr-v305-black25a}. We summarize this pathway as \mbox{$V\rightarrow\boldsymbol{L}\rightarrow A$}, where $\boldsymbol{L}$ denotes an LLM-centric multimodal interface that goes beyond text encoding to integrate visual and linguistic information into representations for action prediction. Thus, both action generation designs retain the language-model backbone as the representational bridge between perception and action.

\textbf{Our approach.}
Unlike LLM-centric VLA, TurboVLA separately encodes visual observations and task instructions and integrates their features through direct bidirectional vision-language interaction. Specifically, the instruction, visual observations, and robot state are encoded as $Z^l=\Phi_l(x)$, $Z_t^v=\Phi_v(\mathcal{O}_t)$, and $Z_t^s=\Phi_s(s_t)$, respectively, where $\Phi_l$ is a lightweight instruction encoder. The resulting vision-language representation is combined with the robot-state features to predict an action chunk:
\begin{equation}
\begin{aligned}
Z_t^{vl} &= \Phi_{vl}(Z_t^v,Z^l),\quad
\hat{\mathbf{A}}_t
&=
D_{\theta}\!\left(
Q_a,\left[Z_t^{vl};Z_t^s\right]
\right).
\end{aligned}
\end{equation}
Here, $\Phi_l$, $\Phi_v$, and $\Phi_s$ denote the instruction, visual, and robot state encoders, respectively, $\Phi_{vl}$ denotes the direct vision-language interaction module, and $D_{\theta}$ denotes the action decoder.

\subsection{Multimodal Feature Encoding}

\textbf{Lightweight instruction encoding.} To reduce the overhead of an LLM-centric execution pathway while retaining simple yet sufficient instruction understanding, \ourmethod separately encodes task instructions and visual observations, as shown in Fig.~\ref{fig:method}(a). Given a task instruction $x$, a lightweight text encoder such as BERT~\citep{devlin2019bert} extracts token-level instruction features:
\begin{equation}
Z^l=P_l\left(f_{\mathrm{text}}(x)\right)\in\mathbb{R}^{N_l\times d},
\end{equation}
where $P_l$ projects the encoder output into the hidden dimension $d$ and $N_l$ is the number of instruction tokens. This lightweight encoding provides a compact semantic representation of the task instruction.

\textbf{Multi-view visual encoding.} For each camera observation $I_t^{(i)}$, the image encoder extracts spatial visual features, which are projected and augmented with positional and camera-view embeddings:
\begin{equation}
Z_t^{v,(i)}=P_v\left(f_{\mathrm{img}}\left(I_t^{(i)}\right)\right)+E_{\mathrm{pos}}^{(i)}+e_{\mathrm{view}}^{(i)}, \qquad
Z_t^v=\left[Z_t^{v,(1)};\ldots;Z_t^{v,(K)}\right].
\end{equation}
Here, $E_{\mathrm{pos}}^{(i)}$ preserves within-view spatial structure and $e_{\mathrm{view}}^{(i)}$ identifies the camera source. Concatenating the $K$ streams retains complementary cues from multiple viewpoints.

\textbf{Robot state encoding.} We encode the current robot state $s_t$ separately, where $f_{\mathrm{state}}$ is a lightweight projection network and $N_s$ is the number of state tokens:
\begin{equation}
Z_t^s=f_{\mathrm{state}}(s_t)\in\mathbb{R}^{N_s\times d},
\end{equation}
These state features are fed directly to the action decoder to condition action prediction on the robot's current configuration.

\subsection{Vision-Language Interaction Module}
Independently encoded visual and textual features do not yet identify which visual content is relevant to the current instruction. While LLM-centric VLAs perform this alignment within a large language backbone, \ourmethod instead uses the simple and efficient vision-language interaction module in Fig.~\ref{fig:method}(b) to directly exchange information between the two streams.

Let $V_t^0=Z_t^v$ and $L_t^0=Z^l$ denote the initial visual and instruction features. The interaction module progressively updates both streams through $N$ bidirectional cross-modal layers:
\begin{equation}
\left(V_t^\ell,L_t^\ell\right)
=
\operatorname{FusionLayer}_{\ell}
\left(V_t^{\ell-1},L_t^{\ell-1}\right),
\qquad
\ell=1,\ldots,N.
\end{equation}
Within each fusion layer, the two streams exchange information through bidirectional cross-attention. We omit the layer index on these projections for notational simplicity:
\begin{equation}
{V}_t^\ell
=
V_t^{\ell-1}
+
\operatorname{Attn}_{v\leftarrow l}
\!\left(V_t^{\ell-1},L_t^{\ell-1}\right),
\qquad
{L}_t^\ell
=
L_t^{\ell-1}
+
\operatorname{Attn}_{l\leftarrow v}
\!\left(L_t^{\ell-1},V_t^{\ell-1}\right).
\end{equation}
Visual-to-instruction attention conditions visual features on task semantics, while instruction-to-visual attention injects scene context into the instruction stream. After the final layer, the updated streams are concatenated as
\begin{equation}
Z_t^{vl}
=
\left[
V_t^{N};
L_t^{N}
\right].
\end{equation}
Through this compact interaction module, information including target objects, attributes, and spatial relations can modulate the relevant visual features, while the instruction representation is simultaneously adapted to the current scene. This simple interaction design efficiently provides task-specific multimodal information for action prediction without relying on the broad generative and reasoning capacity of a large language model.

\subsection{Continuous Action Chunk Prediction}
\label{sec:action_decoder}
We use a lightweight ACT-style transformer decoder~\citep{zhao2023learning} to map the fused multimodal representation and robot state features to a sequence of continuous actions:
\begin{equation}
\hat{\mathbf{A}}_t=D_{\theta}\!\left(Q_a,\left[Z_t^{\mathrm{vl}};Z_t^s\right]\right)\in\mathbb{R}^{H\times d_a},
\end{equation}
where $Q_a=[q_1,\ldots,q_H]$ contains $H$ learnable action queries, $H$ is the prediction horizon, $d_a$ is the action dimension, and $D_\theta$ denotes the action decoder. We train \ourmethod through behavior cloning with an $\ell_1$ loss between the predicted chunk $\hat{\mathbf{A}}_t$ and the expert sequence $\mathbf{A}_t^*=[a_t^*,\ldots,a_{t+H-1}^*]$.

\section{Experiments}
\label{sec:experiments}

We evaluate whether \ourmethod can retain strong language-conditioned manipulation performance while substantially reducing the model scale, inference latency, and memory overhead of LLM-centric VLA policies. Experiments cover single-arm manipulation on LIBERO~\citep{liu2023libero}, bimanual manipulation on RoboTwin 2.0~\citep{chen2025robotwin}, and real-world deployment, followed by ablations of language conditioning and vision-language interaction.

\subsection{Experimental Setup}
\label{sec:experimental_details}

We use DINOv3~\citep{simeoni2025dinov3} as the visual backbone and BERT~\citep{devlin2019bert} as the instruction encoder. Visual and linguistic features are projected to a dimension $d=256$ and processed by $N=6$ bidirectional interaction layers initialized from pretrained Grounding DINO weights~\citep{liu2024grounding}. We use an ACT-style decoder~\citep{zhao2023learning} to predict continuous action chunks from the resulting representations. Training details are provided in Appendix~\ref{app:implementation}.

\textbf{LIBERO} contains four suites, each comprising ten language-conditioned manipulation tasks. We use the modified \texttt{no\_noops} RLDS datasets released with OpenVLA~\citep{kim2024openvla} and jointly train one mixed-suite model with a DINOv3 ViT-B backbone. The model is trained for 40k steps with a batch size of 128. Following the VLA-Adapter rollout protocol~\citep{wang2026vla}, we conduct 50 rollouts per task and report suite-level and average success rates over 2,000 trials.

\textbf{RoboTwin 2.0} comprises 50 language-conditioned bimanual manipulation tasks requiring coordinated dual-arm control. Following prior work~\citep{bi2026motus,yang2026abot,wang2026qwen,wu2026pragmatic}, we jointly train \ourmethod with a DINOv3 ViT-L backbone on 2,500 clean and 25,000 randomized demonstrations. Training lasts 150k steps with a batch size of 512. Following the StarVLA~\citep{community2026starvla} evaluation framework, we conduct 100 rollouts per task in both clean and randomized settings.

\textbf{Real-world evaluation.}
We conducted real-world experiments using an AgileX Piper platform illustrated in Fig.~\ref{fig:realworld}. We consider four representative language-conditioned manipulation tasks: \textit{grab roller}, \textit{move playing card away}, \textit{press stapler}, and \textit{stack three bowls}. Together, these tasks involve accurate object grounding, multi-object manipulation and precise interaction. We initialize the policy from the \ourmethod pretrained on LIBERO and fine-tune it on $4\times65$ teleoperated real-world demonstrations for 12.5k steps. Each task is evaluated over 40 trials. We compare against $\pi_{0.5}$ under the same platform, fine-tuning data, and evaluation protocol.

\subsection{Main Results}
\label{sec:main_results}
Tab.~\ref{tab:libero_main} and Tab.~\ref{tab:robotwin_mixed} present complementary evaluations of \ourmethod on simulation benchmarks. From these results, we draw the following observations.

\begin{table}[t]
\caption{
Comparison on LIBERO. ``Emb. PT.'' denotes additional embodied pretraining on robot data beyond LIBERO. Params denotes total parameters. Latency denotes time from multimodal input to producing an action chunk or an equivalent number of autoregressive action tokens. Both latency and Inference VRAM are measured on a single RTX 4090 with batch size one. For TurboVLA, the reported parameter count corresponds to the DINOv3 ViT-B configuration.
}
\label{tab:libero_main}
\centering
\scriptsize
\setlength{\tabcolsep}{1.5pt}
\renewcommand{\arraystretch}{1.08}

\begin{tabularx}{\textwidth}{
@{}
>{\raggedright\arraybackslash}X                         
>{\centering\arraybackslash}p{0.139\textwidth}          
>{\centering\arraybackslash}p{0.045\textwidth}          
>{\centering\arraybackslash}p{0.060\textwidth}          
>{\centering\arraybackslash}p{0.060\textwidth}          
>{\centering\arraybackslash}p{0.070\textwidth}          
*{5}{>{\centering\arraybackslash}p{0.052\textwidth}}    
@{}
}
\toprule
\multirow{2}{*}{Method}
& \multirow{2}{*}{Reference}
& \multirow{2}{*}{\shortstack{Emb. \\PT.}}
& \multicolumn{3}{c}{Deployment Efficiency}
& \multicolumn{5}{c}{LIBERO Success Rate (\%)} \\
\cmidrule(lr){4-6}
\cmidrule(lr){7-11}
&
&
& \shortstack{Params\\(B)$\downarrow$}
& \shortstack{VRAM\\(GiB)$\downarrow$}
& \shortstack{Latency\\(ms)$\downarrow$}
& Spa.
& Obj.
& Goal
& Long
& Avg.$\uparrow$ \\
\midrule

\multicolumn{11}{l}{
\textit{\textbf{Non-VLA policy baseline}}
} \\
Diffusion Policy~\citep{chi2025diffusion}  & RSS 23
    & \xmark & 0.3 & 1.1 & 924.8
    & 78.3 & 92.5 & 68.3 & 50.5 & 72.4 \\

\midrule
\multicolumn{11}{l}{
\textit{\textbf{Capability-oriented VLAs}}
} \\

OpenVLA~\citep{kim2024openvla} & CoRL 24
    & \cmark & 7.5 & 14.9 & 202.9
    & 84.7 & 88.4 & 79.2 & 53.7 & 76.5 \\

$\pi_0$~\citep{black2024pi_0} & RSS 25
    & \cmark & 3.2 & 7.2 & 219.2
    & 96.8 & 98.8 & 95.8 & 85.2 & 94.2 \\

UniVLA~\citep{bu2025univla} & RSS 25
    & \cmark & 7.6 & 15.0 & 173.8
    & 96.5 & 96.8 & 95.6 & 92.0 & 95.2 \\

$\pi_{0.5}$~\citep{pmlr-v305-black25a} & CoRL 25
    & \cmark & 3.4 & 7.7 & 238.8
    & 98.8 & 98.2 & 98.0 & 92.4 & 96.9 \\

CogVLA~\citep{li2026cogvla} & NeurIPS 25
    & \cmark & 8.3 & 16.1 & 115.5
    & 98.6 & 98.8 & 96.6 & 95.4 & 97.4 \\

Mantis~\citep{yang2026mantis} & CVPR 26
    & \cmark & 4.9 & 7.9 & 198.7
    & 98.8 & 99.2 & 94.4 & 94.2 & 96.7 \\

MM-ACT~\citep{liang2026mm} & CVPR 26
    & \xmark & 8.2 & 16.3 & 723.2
    & 97.8 & 99.4 & 94.8 & 93.0 & 96.3 \\

VLA-JEPA~\citep{sun2026vla} & ECCV 26
    & \cmark & 2.8 & 5.3 & 108.7
    & 96.2 & 99.6 & 97.2 & 95.8 & 97.2 \\

VEGA-3D~\citep{wu2026generation} & ECCV 26
    & \cmark & 9.0 & 16.0 & 546.4
    & 97.4 & 99.4 & 97.0 & 95.2 & 97.3 \\

\midrule
\multicolumn{10}{l}{
\textit{\textbf{Acceleration-oriented VLAs}}
} \\

OpenVLA-OFT~\citep{kim2025fine} & RSS 25
    & \cmark & 7.7 & 15.7 & 112.2
    & 97.6 & 98.4 & 97.9 & 94.5 & 97.1 \\

DDVLA~\citep{liang2025discrete} & ICML 26
    & \cmark & 7.5 & 14.5 & 60.8
    & 97.2 & 99.4 & 96.8 & 92.2 & 96.4 \\

\midrule
\multicolumn{11}{l}{
\textit{\textbf{Lightweight VLAs}}
} \\

SmolVLA~\citep{shukor2025smolvla} & ArXiv 25
    & \xmark & 0.5 & 1.4 & 212.7
    & 90.0 & 96.0 & 92.0 & 71.0 & 87.3 \\

DreamVLA~\citep{zhang2026dreamvla} & NeurIPS 25
    & \xmark & 0.7 & 1.5 & 128.0
    & 97.5 & 94.0 & 89.5 & 89.5 & 92.6 \\

VLA-Adapter~\citep{wang2026vla} & AAAI 26
    & \xmark & 1.5 & 4.3 & 87.3
    & 97.8 & 99.2 & 97.2 & 95.0 & 97.3 \\

Evo-1~\citep{lin2026evo} & CVPR 26
    & \xmark & 0.8 & 1.7 & 137.2
    & 92.7 & 97.7 & 96.3 & 92.3 & 94.8 \\

\midrule
\rowcolor[gray]{0.92}
\textbf{\ourmethod (Ours)} & -
    & \xmark
    & \textbf{0.2}
    & \textbf{0.9}
    & \textbf{31.2}
    & 97.0
    & 100.0
    & 98.2
    & 95.2
    & \textbf{97.6} \\
\bottomrule
\end{tabularx}
\vspace{-2pt}
\end{table}

\begin{table*}[t]
\caption{
Comparison on RoboTwin 2.0.
``Emb. PT.'' denotes additional embodied pretraining on robot data beyond RoboTwin 2.0,
and Params denotes total parameters.
``Clean'' and ``Randomized'' denote the success rates under the clean and randomized evaluation settings, respectively.
For TurboVLA, the reported parameter count corresponds to the DINOv3 ViT-L configuration.
}
\label{tab:robotwin_mixed}
\centering
\scriptsize
\setlength{\tabcolsep}{2pt}
\renewcommand{\arraystretch}{1.08}

\begin{tabularx}{\textwidth}{
@{}
>{\raggedright\arraybackslash}X
>{\centering\arraybackslash}p{0.130\textwidth}
>{\centering\arraybackslash}p{0.05\textwidth}
>{\centering\arraybackslash}p{0.095\textwidth}
>{\centering\arraybackslash}p{0.095\textwidth}
>{\centering\arraybackslash}p{0.09\textwidth}
>{\centering\arraybackslash}p{0.09\textwidth}
>{\centering\arraybackslash}p{0.09\textwidth}
@{}
}
\toprule

\multirow{2}{*}{Method}
& \multirow{2}{*}{Reference}
& \multirow{2}{*}{\shortstack{Emb.\\PT.}}
& \multicolumn{2}{c}{\makebox[0pt][c]{Deployment Efficiency}}
& \multicolumn{3}{c}{RoboTwin 2.0 Success Rate (\%)} \\

\cmidrule(lr){4-5}
\cmidrule(lr){6-8}

&
&
& \shortstack{Params\\(B)$\downarrow$}
& \shortstack{Latency\\(ms)$\downarrow$}
& Clean
& Rand.
& Avg.$\uparrow$ \\

\midrule

$\pi_0$~\citep{black2024pi_0} & RSS 25
    & \cmark & 3.2 & 214.5
    & 65.92 & 58.40 & 62.16 \\

$\pi_{0.5}$~\citep{pmlr-v305-black25a} & CoRL 25
    & \cmark & 3.4 & 246.5
    & 82.74 & 76.76 & 79.75 \\

Motus~\citep{bi2026motus} & CVPR 26
    & \cmark & 8.7 & 2070.4
    & 88.66 & 87.02 & 87.84 \\

X-VLA~\citep{zheng2026x} & ICLR 26
    & \cmark & 0.9 & 87.3
    & 72.80 & 72.84 & 72.82 \\

GigaWorld-Policy~\citep{ye2026gigaworld} & ArXiv 26
    & \cmark & 12.4 & 654.2
    & 86.36 & 85.04 & 85.70 \\

ABot-M0~\citep{yang2026abot} & ArXiv 26
    & \cmark & 5.9 & 172.6
    & 86.06 & 85.08 & 85.57 \\

Qwen-VLA~\citep{wang2026qwen} & ArXiv 26
    & \cmark & -- & --
    & 86.10 & 87.20 & 86.65 \\

LingBot-VLA~\citep{wu2026pragmatic} & ArXiv 26
    & \cmark & 4.2 & 164.4
    & 88.56 & 86.68 & 87.62 \\

\rowcolor[gray]{0.92}
\textbf{\ourmethod (Ours)} & -
    & \xmark & \textbf{0.4} & \textbf{43.4}
    & 88.02 & 88.10 & \textbf{88.06} \\

\bottomrule
\end{tabularx}

\vspace{-2pt}
\end{table*}

\textbf{1) Lightweight instruction encoding and direct vision-language interaction enable competitive and efficient manipulation.}
As summarized in Tab.~\ref{tab:libero_main}, \ourmethod achieves 97.6\% average success with 0.2B parameters, 31.2\,ms inference latency, and 0.9\,GiB inference VRAM on LIBERO~\citep{liu2023libero}. Compared with capability-oriented VLAs, it achieves higher average success than $\pi_{0.5}$ at 96.9\% while using approximately 6\% of its parameters and providing $8\times$ faster inference. The favorable trade-off also holds against acceleration-oriented and lightweight VLAs. OpenVLA-OFT~\citep{kim2025fine} and DDVLA~\citep{liang2025discrete} achieve inference latencies of 112.2\,ms and 60.8\,ms, respectively, while \ourmethod achieves both higher average success and lower latency without dedicated acceleration mechanisms. Among lightweight VLAs, \ourmethod exceeds the performance of Evo-1~\citep{lin2026evo} and SmolVLA~\citep{shukor2025smolvla} with fewer parameters and faster inference. Together, these comparisons suggest that lightweight instruction encoding can provide task-relevant semantics for direct vision-language interaction, supporting competitive execution-level manipulation without the computational and memory costs of an LLM-centric policy.

\textbf{2) This simple and lightweight design extends to challenging bimanual multi-task control.}
Beyond single-arm manipulation, we evaluate whether TurboVLA can support coordinated bimanual control on RoboTwin 2.0~\citep{chen2025robotwin}. Recent models such as Qwen-VLA~\citep{wang2026qwen} and LingBot-VLA~\citep{wu2026pragmatic} achieve strong performance by benefiting from large-scale embodied pretraining. Nevertheless, \ourmethod achieves a higher average success rate than both models, reaching 88.06\% with only 0.4B parameters and 43.4\,ms inference latency as shown in Tab.~\ref{tab:robotwin_mixed}. It also matches the performance of Motus~\citep{bi2026motus} and ABot-M0~\citep{yang2026abot}, despite their substantially larger model sizes. Compared with LingBot-VLA, \ourmethod reduces inference latency from 164.4\,ms to 43.4\,ms, providing approximately $3.8\times$ faster inference. It further outperforms $\pi_{0.5}$~\citep{pmlr-v305-black25a} while providing approximately $5.7\times$ faster inference. These results show that the simple TurboVLA design remains highly effective for challenging bimanual manipulation with substantially less computation and no embodied pretraining.

\begin{figure*}[t]
    \centering
    \includegraphics[width=\textwidth]{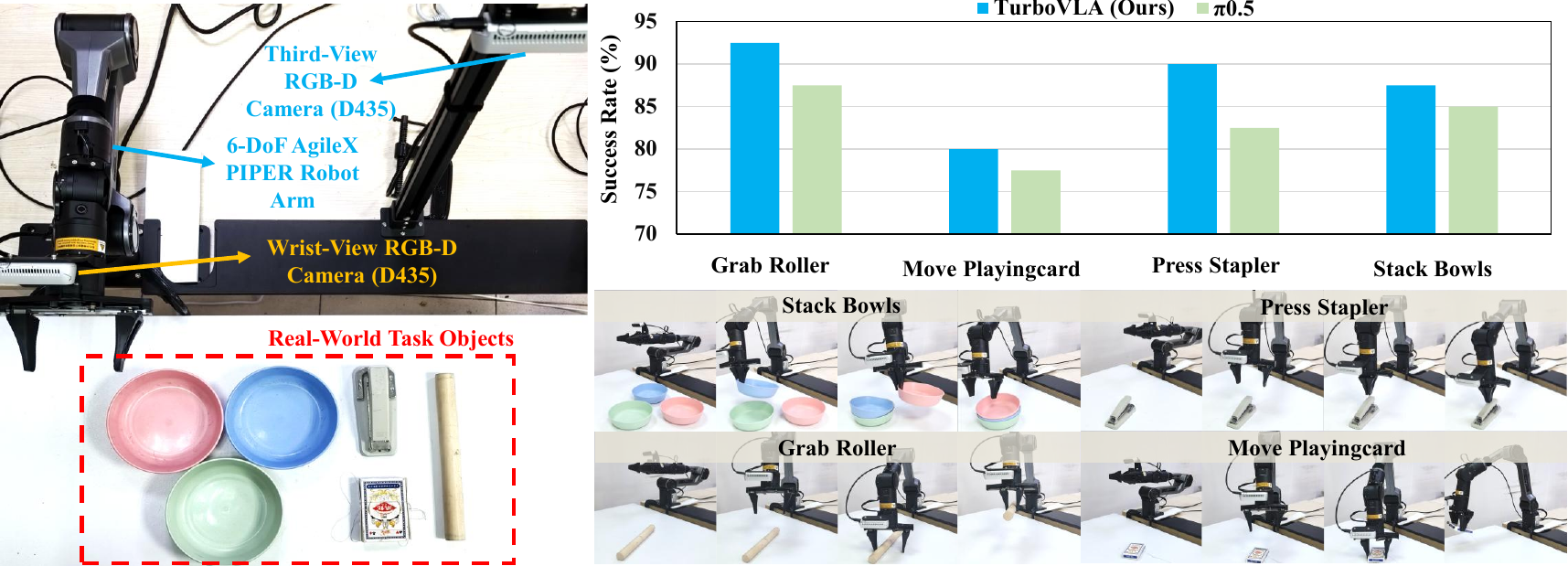}
    \caption{
    Real-world evaluation on the AgileX Piper platform. Left: our single-arm setup with a wrist-view RGB-D camera and a third-view RGB-D camera, together with the objects used in the four tasks. Right top: success-rate comparison between TurboVLA and $\pi_{0.5}$ on four real-world manipulation tasks. Right bottom: qualitative execution examples of TurboVLA.
    }
    \label{fig:realworld}
\end{figure*}

\textbf{3) Architectural efficiency translates into practical deployability.}
Compared with LLM-centric policies such as $\pi_{0.5}$, \ourmethod substantially reduces model size, inference latency, and GPU memory requirements, as summarized in Tab.~\ref{tab:libero_main}. This efficiency is accompanied by competitive real-world performance: on the AgileX Piper platform, \ourmethod achieves 87.5\% average success across four manipulation tasks, performing comparably to $\pi_{0.5}$, as shown in Fig.~\ref{fig:realworld}. These results demonstrate that TurboVLA combines competitive physical manipulation with low inference cost, supporting its suitability for real-world execution-level control.

\subsection{Ablation Study}
\label{sec:ablation}

We conduct ablations on LIBERO to examine the role of semantic language conditioning, the choice of instruction encoder, and the design of vision-language interaction. Additional ablations on LIBERO and RoboTwin 2.0 are reported in Appendix~\ref{app:additional_ablations}.

\begin{figure*}[t]
\centering
\setlength{\ablationleftwidth}{0.52\textwidth}
\setlength{\ablationrightwidth}{0.46\textwidth}

\begin{lrbox}{\ablationleftbox}
\begin{minipage}[t]{\ablationleftwidth}
\vspace{0pt}
\centering

\captionof{table}{Effect of language conditioning.}
\label{tab:ablation_language}

\begingroup
\scriptsize
\setlength{\tabcolsep}{2pt}
\renewcommand{\arraystretch}{1.08}

\begin{tabularx}{0.9\linewidth}
{@{}>{\raggedright\arraybackslash}Xccccc@{}}
\toprule
Condition & Spa. & Obj. & Goal & Long & Avg. \\
\midrule
w/o Language
    & 87.4 & 99.8 & 8.6 & 78.0 & 68.5 \\
Task-ID Embedding
    & 96.8 & 99.6 & 97.2 & 92.6 & 96.6 \\
\rowcolor[gray]{0.92}
Semantic Instruction
    & 97.0 & 100.0 & 98.2 & 95.2 & \textbf{97.6} \\
\bottomrule
\end{tabularx}
\par
\endgroup

\captionof{table}{Effect of the instruction encoder.}
\label{tab:ablation_text_encoder}

\begingroup
\scriptsize
\setlength{\tabcolsep}{2.0pt}
\renewcommand{\arraystretch}{1.08}

\begin{tabularx}{0.9\linewidth}
{@{}>{\raggedright\arraybackslash}Xcccccc@{}}
\toprule
Text Encoder
    & \shortstack{Overall\\Params (M)}
    & Spa. & Obj. & Goal & Long & Avg. \\
\midrule
SigLIP-Base
    & 216.9 & 98.4 & 98.8 & 96.0 & 95.8 & 97.3 \\
T5-Base
    & 216.2 & 98.8 & 99.4 & 97.0 & 95.2 & \textbf{97.6} \\
\rowcolor[gray]{0.92}
BERT
    & 216.1 & 97.0 & 100.0 & 98.2 & 95.2 & \textbf{97.6} \\
\bottomrule
\end{tabularx}
\par
\endgroup

\captionof{table}{Effect of vision-language interaction.}
\label{tab:ablation_fusion}

\begingroup
\scriptsize
\setlength{\tabcolsep}{2pt}
\renewcommand{\arraystretch}{1.08}

\begin{tabularx}{0.9\linewidth}
{@{}>{\raggedright\arraybackslash}Xccccc@{}}
\toprule
Interaction Design & Spa. & Obj. & Goal & Long & Avg. \\
\midrule
w/o Interaction
    & 98.8 & 99.8 & 95.6 & 90.8 & 96.3 \\
Language Queries Visual
    & 96.8 & 99.6 & 97.0 & 93.8 & 96.8 \\
Visual Queries Language
    & 96.8 & 99.6 & 98.6 & 93.6 & 97.2 \\
\rowcolor[gray]{0.92}
Bidirectional Interaction
    & 97.0 & 100.0 & 98.2 & 95.2 & \textbf{97.6} \\
\bottomrule
\end{tabularx}
\par
\endgroup

\end{minipage}
\end{lrbox}

\setlength{\ablationcolheight}{%
    \dimexpr\ht\ablationleftbox+\dp\ablationleftbox\relax
}

\begin{lrbox}{\ablationcaptionbox}
\begin{minipage}{\ablationrightwidth}
\captionof{figure}{
Vision-language interaction variants.
}
\label{fig:ablation_interaction}
\end{minipage}
\end{lrbox}

\setlength{\ablationimageheight}{%
    \dimexpr
    \ablationcolheight
    -\ht\ablationcaptionbox
    -\dp\ablationcaptionbox
    \relax
}

\noindent
\usebox{\ablationleftbox}\hfill%
\raisebox{-4pt}{%
\begin{minipage}[t][\ablationcolheight][t]{\ablationrightwidth}
\vspace{0pt}
\centering

\includegraphics[
    trim=0 0 0 0,
    clip,
    width=\linewidth,
    height=\ablationimageheight,
    keepaspectratio
]{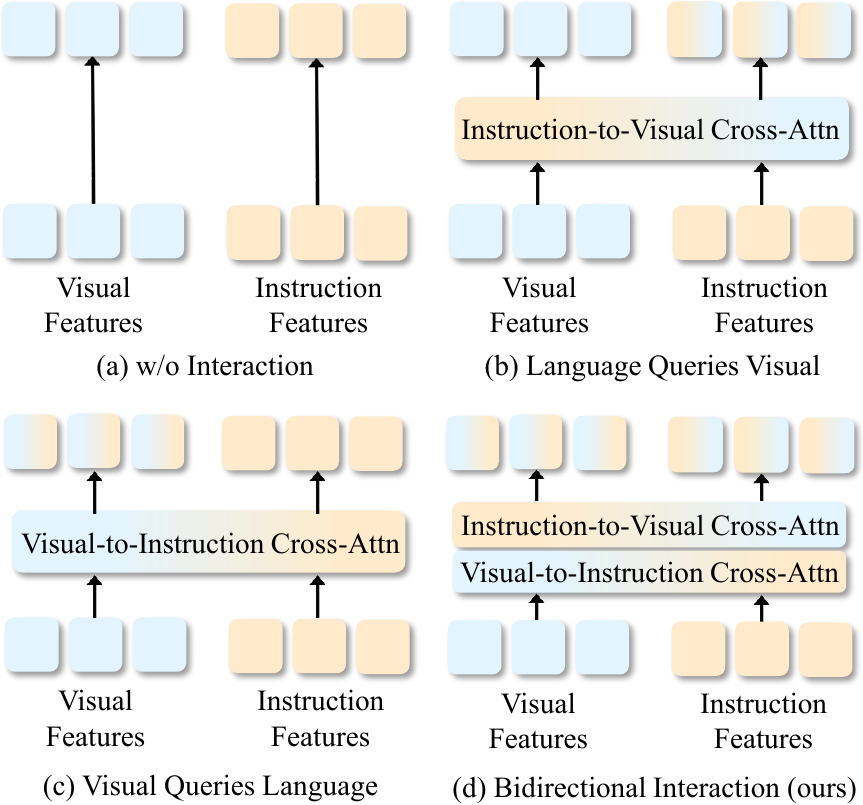}
\par

\vfill
\nointerlineskip
\noindent\usebox{\ablationcaptionbox}
\par

\end{minipage}%
}

\end{figure*}

\textbf{Semantic language conditioning.}
We first study the role of language itself. Tab.~\ref{tab:ablation_language} shows that removing language reduces the average success rate from 97.6\% to 68.5\%, with the largest drop on LIBERO-Goal. This degradation indicates that
visual observations alone are insufficient when the same scene can correspond to different instructed goals. Learned task-ID embeddings recover average success to 96.6\%, but remain 1.0\% below semantic instructions. This gap suggests that semantic instruction features provide more effective conditioning than task identity alone. 

\textbf{Instruction encoder.}
Then, we examine whether the proposed architecture depends on a specific text backbone. As shown in Tab.~\ref{tab:ablation_text_encoder}, under otherwise identical settings, T5-Base~\citep{raffel2020exploring} achieves a competitive 97.6\% average success rate, while the SigLIP~\citep{zhai2023sigmoid} text encoder reaches 97.3\%. These results support the effectiveness of lightweight instruction encoding for execution-level control and show that TurboVLA is robust to the choice of text backbone.

\textbf{Vision-language interaction design.}
Having established the importance of semantic instruction features, we next study how visual and language features should interact before action decoding. As illustrated in Fig.~\ref{fig:ablation_interaction}, we compare no interaction, two asymmetric cross-attention variants, and the proposed bidirectional interaction, while keeping all other architecture and training settings unchanged. As shown in Tab.~\ref{tab:ablation_fusion}, direct concatenation achieves 96.3\% average success, while the two one-way cross-attention variants improve it to 96.8\% and 97.2\%. These gains suggest that explicit interaction before action decoding is beneficial even when the decoder already has access to both modalities. Bidirectional interaction performs best at 97.6\%, indicating that allowing visual and instruction features to condition each other benefits action prediction.

Overall, these ablations support the proposed direct \mbox{$\boldsymbol{V}+\boldsymbol{L}\rightarrow A$} pathway, showing that lightweight semantic instruction encoding and bidirectional vision-language interaction together enable effective execution-level control without an LLM-centric policy.

\subsection{Runtime Acceleration of TurboVLA}
\label{sec:cuda_graph_compatibility}

\begingroup
\setlength{\columnsep}{8pt}
\setlength{\intextsep}{1pt}
\begin{wraptable}{r}{0.35\linewidth}
\centering
\caption{Runtime acceleration of TurboVLA.}
\label{tab:cuda_graph_compatibility}
{\scriptsize
\setlength{\tabcolsep}{3pt}
\renewcommand{\arraystretch}{1.12}
\begin{tabular}{@{}lcc@{}}
\toprule
Method & No Graph & CUDA Graph\\
\midrule
TurboVLA
& 31.2\,ms & 11.3\,ms \\
\rowcolor[gray]{0.92}
\textbf{TurboVLA + Opt.}
& \textbf{30.3\,ms} & \textbf{7.42\,ms} \\
\bottomrule
\end{tabular}\par}
\end{wraptable}

Building on the low inference latency of TurboVLA, we further integrate CUDA Graph replay with the policy and apply several runtime optimizations (Opt.). As shown in Tab.~\ref{tab:cuda_graph_compatibility}, the resulting implementation remarkably reduces the policy inference latency to 7.42\,ms on an RTX 4090. These refinements target data conversion, memory allocation, and unused intermediate outputs, while preserving the full visual and language encoders and the original action horizon. This latency corresponds to a policy inference rate above 130\,Hz, suggesting that policy inference itself is unlikely to be the bottleneck for real-time manipulation. In this regime, the practical control frequency is more likely to be limited by sensing, communication, and actuator dynamics than by the execution policy, further demonstrating the deployment potential of TurboVLA in latency-sensitive robotic systems.

\par\endgroup

\section{Conclusion}
In this paper, we propose TurboVLA, a simple and efficient \mbox{$\boldsymbol{V}+\boldsymbol{L}\rightarrow A$} model that combines lightweight instruction encoding with direct bidirectional vision-language interaction for execution-level manipulation. Experiments on LIBERO, RoboTwin 2.0, and real-world manipulation tasks show that TurboVLA preserves task-conditioned manipulation capability while significantly reducing model size, inference latency, and memory consumption. Our results suggest that execution-level control does not necessarily require a general-purpose LLM as the central interface between perception and action, offering insight into how vision and language can be organized for efficient robotic control.

\textbf{Limitations and future work.}
Nevertheless, TurboVLA currently focuses on concrete execution-level manipulations and is not specifically designed for open-ended commands such as ``clean the room'', where the robot needs to rely on high-level reasoning and task decomposition to identify relevant subgoals and organize their execution. Future work will explore extending TurboVLA to high-level and long-horizon tasks, aiming to achieve more intelligent manipulation while retaining its efficient execution pathway. It will also broaden evaluation beyond simulation benchmarks and tabletop scenarios to more diverse real-world settings, including cluttered household environments, office environments, industrial workspaces, and other robot embodiments.

\bibliography{iclr2027_conference}
\bibliographystyle{iclr2027_conference}

\newpage
\appendix

\section{Implementation and Training Details}
\label{app:implementation}

\subsection{Architecture and Initialization}
\label{app:experimental_details}

\ourmethod uses DINOv3 ViT-B/16 for LIBERO and ViT-L/16 for RoboTwin 2.0~\citep{simeoni2025dinov3}, together with BERT-base-uncased~\citep{devlin2019bert} for instruction encoding. The vision-language interaction module is initialized with pretrained feature-enhancement weights from the Grounding DINO Swin-T OGC checkpoint~\citep{liu2024grounding}. Both configurations use a shared feature dimension of $d=256$, six bidirectional interaction layers, and a three-layer ACT-style action decoder~\citep{zhao2023learning}. The current robot state is projected into $N_s=2$ tokens, which are concatenated with the fused vision--language features to form the decoder memory. Table~\ref{tab:app_architecture} summarizes the architecture configurations.

\subsection{Data Processing}
\label{app:data_processing}

\textbf{LIBERO.}
We jointly train on the four modified \texttt{no\_noops} RLDS suites released with OpenVLA~\citep{kim2024openvla}. Each sample contains a primary-view RGB image, a wrist-view RGB image, a task instruction, and an 8-dimensional robot state. We preserve the original $256\times256$ image resolution without resizing or cropping and apply the pixel normalization used by the pretrained visual encoder. Robot states are standardized using statistics computed over the combined training data. Each action is 7-dimensional: the six continuous arm-action coordinates are min--max normalized to $[-1,1]$ and clipped to this range, while the gripper command retains its original binary encoding in $\{-1,+1\}$. Each target chunk contains $H=12$ consecutive actions starting at the observation timestep. At episode boundaries, incomplete chunks are padded by repeating the final action, and padded entries are excluded from the loss.

\textbf{RoboTwin 2.0.}
Training combines 2,500 clean and 25,000 randomized demonstrations across 50 tasks. We sample tasks uniformly, with randomized and clean demonstrations sampled at a ratio of $10:1$ within each task. Each sample contains a head-view RGB image, two wrist-view RGB images, a task instruction, and the current bimanual state. All three images have a resolution of $224\times224$. Both states and actions are 14-dimensional, comprising the left-arm joint positions, right-arm joint positions, and two gripper coordinates. Continuous joint coordinates are min--max normalized to $[-1,1]$, while gripper coordinates are binarized using a threshold of 0.49. Each observation is paired with a target chunk of $H=50$ consecutive absolute joint-position and gripper actions.

\subsection{Training Configurations}
\label{app:training_configurations}

We train both policies using AdamW to minimize an $\ell_1$ behavior-cloning loss. BERT remains frozen, while the visual encoder and all remaining policy modules are optimized for action prediction. Training runs for 40,000 steps with an effective batch size of 128 on 4 NVIDIA RTX 4090 GPUs for LIBERO, and 150,000 steps with an effective batch size of 512 on 8 NVIDIA H100 GPUs for RoboTwin 2.0. Both benchmarks use linear learning-rate warm-up followed by a constant learning rate. We maintain an exponential moving average (EMA) of the trainable parameters with a decay factor of 0.999. Table~\ref{tab:app_training} summarizes the training configurations.

\begin{table}[!t]
    \centering
    \small
    \caption{Architecture configurations of TurboVLA.}
    \label{tab:app_architecture}
    \begin{tabular}{lc}
        \toprule
        Configuration & Value \\
        \midrule
        \multicolumn{2}{l}{\textit{Modality representations}} \\
        Shared feature dimension & 256 \\
        Number of state tokens & 2 \\
        \midrule
        \multicolumn{2}{l}{\textit{vision-language interaction}} \\
        Number of layers & 6 \\
        Attention heads & 4 \\
        Internal dimension & 1,024 \\
        Attention dropout & 0 \\
        Stochastic-depth rate & 0.1 \\
        \midrule
        \multicolumn{2}{l}{\textit{Action decoder}} \\
        Number of layers & 3 \\
        Attention heads & 8 \\
        Feed-forward dimension & 2,048 \\
        Output MLP hidden dimension & 512 \\
        Dropout & 0.1 \\
        \bottomrule
    \end{tabular}
\end{table}

\begin{table}[!t]
    \centering
    \small
    \caption{Training configurations of TurboVLA.}
    \label{tab:app_training}
    \begin{tabular}{lcc}
        \toprule
        Configuration & LIBERO & RoboTwin 2.0 \\
        \midrule
        Optimizer & AdamW & AdamW \\
        Peak learning rate
            & $5\times10^{-5}$ & $5\times10^{-5}$ \\
        AdamW $(\beta_1,\beta_2)$
            & $(0.9,0.95)$ & $(0.9,0.95)$ \\
        AdamW $\epsilon$ & $10^{-8}$ & $10^{-8}$ \\
        Weight decay & $10^{-10}$ & $10^{-10}$ \\
        Training steps & 40,000 & 150,000 \\
        Warm-up steps & 10,000 & 1,000 \\
        Effective batch size & 128 & 512 \\
        Gradient clipping norm & 1.0 & 1.0 \\
        EMA decay & 0.999 & 0.999 \\
        \bottomrule
    \end{tabular}
\end{table}

\section{Evaluation Protocols and Efficiency Measurement}
\label{app:evaluation_protocols}

\subsection{Evaluation Protocols}

\textbf{LIBERO.}
We evaluate the jointly trained policy on the Spatial, Object, Goal, and Long suites following the VLA-Adapter rollout protocol~\citep{wang2026vla}. Each suite contains ten tasks, with 50 trials per task, yielding 2,000 evaluation trials in total. We use benchmark-provided initial states and a fixed evaluation seed, with ten settling steps at the start of each episode. Episode lengths are limited to 220, 280, 300, and 520 control steps for Spatial, Object, Goal, and Long, respectively, excluding the settling steps. The policy predicts 12-step action chunks and executes each chunk before replanning. We report the success rate for each suite and the average across the four suites.

\textbf{RoboTwin 2.0.}
We evaluate the jointly trained policy on all 50 tasks under clean and randomized conditions using the StarVLA evaluation interface~\citep{community2026starvla}. We conduct 100 trials per task under each condition, yielding 5,000 trials per setting, and report success rates for both settings and their average. Evaluation uses unseen task instructions.

\textbf{Real-world evaluation.}
We evaluate \ourmethod on the AgileX Piper platform across four language-conditioned manipulation tasks, with 40 trials per task and 160 trials per policy. Before each trial, the robot is automatically reset to a fixed home configuration, and task objects are randomly positioned within a $50\times50$\,cm region on the tabletop. Each trial has a time limit of 30\,s, during which the policy may retry unsuccessful attempts. A trial ends upon task completion, expiration of the time limit, or an automatic safety stop triggered by a workspace-boundary violation. We compare against $\pi_{0.5}$ using the same robotic platform, fine-tuning data, and evaluation protocol, and report per-task success rates and their average. For \textit{grab roller}, success requires grasping the target object and lifting it clear of the tabletop. For \textit{move playing card away}, the robot must grasp and lift the card, then move it to the right. For \textit{press stapler}, the robot must press the stapler down with its gripper closed. For \textit{stack three bowls}, all three bowls must remain stably stacked after the gripper releases the final bowl. A trial is successful if the corresponding criterion is satisfied within the time limit.

\subsection{Efficiency Measurement}
\label{app:efficiency}

We evaluate all models on a single NVIDIA RTX 4090 with batch size one, using official implementations in evaluation mode with gradients disabled. Within each benchmark, all models use the same action horizon: 12 for LIBERO and 50 for RoboTwin 2.0. Autoregressive policies generate the action tokens required for the specified horizon.

\textbf{Latency.}
We report steady-state local policy inference latency, measured from GPU-resident observations to a complete action chunk on the GPU. This includes visual and language encoding, action generation, and internal action postprocessing, while excluding input preparation and transfer, communication, and environment execution. We report the median latency over 50 synchronized inference runs after ten warm-up iterations, with initial compilation excluded.

\textbf{GPU memory.}
Inference VRAM is the difference between the peak \texttt{nvidia-smi} reading during model initialization, warm-up, and timed inference and the reading on an idle GPU before the run. Memory usage is expressed in GiB ($1024^3$ bytes). Algorithm~\ref{alg:efficiency} summarizes the measurement procedure.

\textbf{Parameter count.}
We count all parameters in the complete inference model, including both trainable and frozen parameters. Optimizer states are excluded.

\begin{algorithm}[t]
\caption{Latency and GPU memory measurement}
\label{alg:efficiency}
\begin{algorithmic}[1]
\Require Model configuration $c$, observation specification $\omega$,
         action horizon $H$
\Ensure Latency statistics and inference VRAM

\State Record idle GPU memory $M_{\mathrm{idle}}$ with \texttt{nvidia-smi}
\State Start periodic GPU memory sampling $\{M_t\}$
\State Launch benchmark process; initialize policy $\pi$
       and GPU-ready observation $o$ from $c,\omega$
\State Set $\pi$ to evaluation mode with gradients disabled
\State Complete initial compilation, if required

\For{$i = 1,\ldots,10$}
    \State $A \gets \pi(o;H)$
    \State Synchronize GPU execution
    \State Release $A$
\EndFor

\State $\mathcal{T} \gets [\,]$
\For{$i = 1,\ldots,50$}
    \State Synchronize GPU execution
    \State $t_0 \gets \Call{Time}{}$
    \State $A \gets \pi(o;H)$
        \Comment{Complete action chunk on GPU}
    \State Synchronize GPU execution
    \State Append $\Call{Time}{}-t_0$ to $\mathcal{T}$
    \State Release $A$
\EndFor

\State Wait for the benchmark process to exit; stop memory sampling
\State $M_{\mathrm{VRAM}} \gets \max_t M_t-M_{\mathrm{idle}}$
\State \Return $\operatorname{median}(\mathcal{T})$,
       $\operatorname{P90}(\mathcal{T})$,
       $\operatorname{mean}(\mathcal{T})$,
       $M_{\mathrm{VRAM}}$
\end{algorithmic}
\end{algorithm}

\section{Additional Results and Ablations}
\label{app:additional_results}

\subsection{Additional Ablation Studies}
\label{app:additional_ablations}

We report additional ablations on LIBERO and RoboTwin 2.0. LIBERO results examine interaction depth and action horizon; RoboTwin 2.0 results examine cross-modal interaction and the instruction encoder.

\begingroup
\setlength{\columnsep}{16pt}
\setlength{\intextsep}{0pt}
\begin{wraptable}{r}{0.49\linewidth}
\centering
\captionsetup{width=\linewidth,justification=centering,
singlelinecheck=true}
\caption{Depth ablation on LIBERO.}
\label{tab:ablation_depth}
\vspace{2pt}
{\scriptsize
\setlength{\tabcolsep}{2pt}
\renewcommand{\arraystretch}{1.08}
\begin{tabularx}{\linewidth}{@{}c*{6}{>{\centering\arraybackslash}X}@{}}
\toprule
$N$ & \shortstack{Params\\(M)} & Spa. & Obj. & Goal & Long & Avg. \\
\midrule
2 & 206.6 & 98.0 & 99.6 & 90.4 & 93.4 & 95.4 \\
4 & 211.3 & 98.0 & 99.2 & 96.0 & 94.4 & 96.9 \\
\rowcolor[gray]{0.92}
6 & 216.1 & 97.0 & 100.0 & 98.2 & 95.2 & \textbf{97.6} \\
\bottomrule
\end{tabularx}\par}
\end{wraptable}

\noindent\textbf{Interaction depth on LIBERO.}
We first examine how the number of vision-language interaction layers affects performance, keeping the remaining architecture fixed. As shown in Tab.~\ref{tab:ablation_depth}, average success increases from 95.4\% at $N=2$ to 96.9\% at $N=4$ and 97.6\% at $N=6$. The improvement is most pronounced on LIBERO-Goal, where success rises from 90.4\% to 98.2\%, whereas the Object suite remains near saturation across all three depths. These results suggest that additional interaction layers are particularly useful for tasks requiring stronger instruction-conditioned visual grounding. We use $N=6$, which achieves the highest average success among the evaluated depths.
\par\endgroup

\par\medskip\noindent
\begin{minipage}[t]{0.49\linewidth}
\vspace{0pt}
\textbf{Action horizon on LIBERO.}
We next vary the predicted action horizon $H$ while keeping the remaining architecture fixed. As shown in Fig.~\ref{fig:ablation_horizon}, average success increases from 96.8\% at $H=8$ to 97.3\% at $H=10$ and 97.6\% at $H=12$, before decreasing to 96.7\% at $H=15$. In the LIBERO rollout, the entire predicted chunk is executed before the policy receives a new observation. A longer horizon therefore extends the period of open-loop execution as well as the prediction window. The decline at $H=15$ indicates that increasing the chunk length does not necessarily improve task performance under this execution protocol. We select $H=12$ as the best of the evaluated settings.
\end{minipage}\hfill
\begin{minipage}[t]{0.49\linewidth}
\vspace{0pt}
\centering
\captionsetup{width=0.96\linewidth,justification=raggedright,
singlelinecheck=false}
\includegraphics[width=\linewidth]{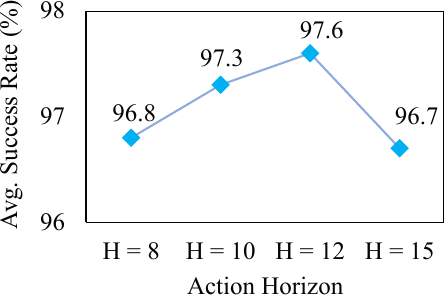}
\captionof{figure}{Action-horizon ablation on LIBERO.}
\label{fig:ablation_horizon}
\end{minipage}

\par\medskip

\smallskip\noindent\textbf{Vision-language interaction on RoboTwin 2.0.}
We repeat the interaction-design comparison from Sec.~\ref{sec:ablation} on RoboTwin 2.0 to test whether its effect extends to bimanual manipulation. As shown in Tab.~\ref{tab:robotwin_interaction}, removing interaction yields 82.58\% average success, compared with 88.06\% for bidirectional interaction. Both one-way variants recover much of this gap: language queries visual features achieve 86.62\%, while visual features querying language achieve 87.08\%. Bidirectional interaction performs best under both clean and randomized conditions, reaching 88.02\% and 88.10\%, respectively. The consistent advantage over the one-way variants supports updating both feature streams before action prediction, as observed on LIBERO.

\smallskip\noindent\textbf{Instruction encoder on RoboTwin 2.0.}
Finally, we replace BERT with T5-Base or SigLIP-Base while retaining the same policy design. As shown in Tab.~\ref{tab:robotwin_text_encoder}, T5-Base achieves 87.39\% average success, compared with 88.06\% for BERT, while SigLIP-Base reaches 86.05\%. T5-Base matches BERT under the clean condition (88.04\% versus 88.02\%) but performs less well under randomization (86.74\% versus 88.10\%). Thus, the architecture remains effective with different instruction encoders, although encoder choice affects performance under the evaluated randomized conditions.

\par\medskip\noindent
\begin{minipage}[t]{0.49\linewidth}
\vspace{0pt}
\centering
\captionof{table}{Effect of vision-language interaction on RoboTwin 2.0.}
\label{tab:robotwin_interaction}
\vspace{2pt}
{\scriptsize
\setlength{\tabcolsep}{2pt}
\renewcommand{\arraystretch}{1.08}
\begin{tabularx}{\linewidth}{@{}>{\raggedright\arraybackslash}p{0.50\linewidth}*{3}{>{\centering\arraybackslash}X}@{}}
\toprule
Interaction Design & Clean & Rand. & Avg. \\
\midrule
w/o Interaction & 80.14 & 85.02 & 82.58 \\
Language Queries Visual & 86.28 & 86.96 & 86.62 \\
Visual Queries Language & 87.16 & 87.00 & 87.08 \\
\rowcolor[gray]{0.92}
Bidirectional Interaction & 88.02 & 88.10 & \textbf{88.06} \\
\bottomrule
\end{tabularx}\par}
\end{minipage}\hfill
\begin{minipage}[t]{0.49\linewidth}
\vspace{0pt}
\centering
\captionof{table}{Effect of the instruction encoder on RoboTwin 2.0.}
\label{tab:robotwin_text_encoder}
\vspace{2pt}
{\scriptsize
\setlength{\tabcolsep}{2pt}
\renewcommand{\arraystretch}{1.36}
\begin{tabularx}{\linewidth}{@{}>{\raggedright\arraybackslash}p{0.50\linewidth}*{3}{>{\centering\arraybackslash}X}@{}}
\toprule
Text Encoder & Clean & Rand. & Avg. \\
\midrule
SigLIP-Base & 85.68 & 86.42 & 86.05 \\
T5-Base & 88.04 & 86.74 & 87.39 \\
\rowcolor[gray]{0.92}
BERT & 88.02 & 88.10 & \textbf{88.06} \\
\bottomrule
\end{tabularx}\par}
\end{minipage}
\par

\subsection{Per-Task Results on RoboTwin 2.0}
\label{app:robotwin_per_task}

Table~\ref{tab:robotwin_per_task} reports the success rates for TurboVLA on all 50 RoboTwin 2.0 tasks under clean and randomized conditions,
with 100 evaluation trials per task in each condition.
The average success rates are 88.02\% and 88.10\%, respectively.

\begin{table}[t]
\centering
\caption{Per-task success rates (\%) of TurboVLA on RoboTwin 2.0.
Both conditions use the EMA checkpoint at 150k training steps
with an action horizon of 50.}
\label{tab:robotwin_per_task}
\footnotesize
\setlength{\tabcolsep}{3pt}
\renewcommand{\arraystretch}{1.05}
\begin{tabular}{@{}lrr@{\hspace{12pt}}lrr@{}}
\toprule
Task & Clean & Rand. & Task & Clean & Rand. \\
\midrule
Adjust Bottle & 100 & 99 & Place Can Basket & 72 & 77 \\
Beat Block Hammer & 100 & 94 & Place Cans Plasticbox & 100 & 99 \\
Blocks Ranking RGB & 89 & 88 & Place Container Plate & 98 & 97 \\
Blocks Ranking Size & 74 & 78 & Place Dual Shoes & 76 & 82 \\
Click Alarmclock & 64 & 73 & Place Empty Cup & 100 & 100 \\
Click Bell & 97 & 98 & Place Fan & 99 & 97 \\
Dump Bin Bigbin & 99 & 99 & Place Mouse Pad & 85 & 79 \\
Grab Roller & 99 & 100 & Place Object Basket & 86 & 78 \\
Handover Block & 86 & 78 & Place Object Scale & 93 & 94 \\
Handover Mic & 92 & 95 & Place Object Stand & 93 & 97 \\
Hanging Mug & 61 & 60 & Place Phone Stand & 77 & 95 \\
Lift Pot & 100 & 99 & Place Shoe & 100 & 100 \\
Move Can Pot & 93 & 89 & Press Stapler & 96 & 97 \\
Move Pillbottle Pad & 97 & 96 & Put Bottles Dustbin & 68 & 86 \\
Move Playingcard Away & 95 & 99 & Put Object Cabinet & 70 & 56 \\
Move Stapler Pad & 72 & 65 & Rotate QR Code & 94 & 95 \\
Open Laptop & 98 & 99 & Scan Object & 93 & 93 \\
Open Microwave & 80 & 90 & Shake Bottle & 99 & 99 \\
Pick Diverse Bottles & 81 & 83 & Shake Bottle Horizontally & 100 & 99 \\
Pick Dual Bottles & 87 & 89 & Stack Blocks Three & 87 & 76 \\
Place A2B Left & 79 & 73 & Stack Blocks Two & 98 & 98 \\
Place A2B Right & 72 & 58 & Stack Bowls Three & 89 & 87 \\
Place Bread Basket & 85 & 81 & Stack Bowls Two & 99 & 100 \\
Place Bread Skillet & 88 & 87 & Stamp Seal & 82 & 85 \\
Place Burger Fries & 98 & 100 & Turn Switch & 61 & 69 \\
\midrule
\textbf{Average} & & & & \textbf{88.02} & \textbf{88.10} \\
\bottomrule
\end{tabular}
\end{table}

\subsection{Per-Task Results on LIBERO}
\label{app:libero_per_task}

Table~\ref{tab:libero_per_task} reports the success rates for TurboVLA on
the 40 subtasks from the LIBERO Object, Goal, Spatial, and Long suites.
Each suite contains ten tasks indexed by task ID from 0 to 9, with 50
evaluation trials per task. The final column reports the average success
rate over the ten tasks in each suite.

TurboVLA achieves average success rates of 100.0\%, 98.2\%, 97.0\%,
and 95.2\% on the Object, Goal, Spatial, and Long suites, respectively.
Across all four suites, TurboVLA obtains 1,952 successes out of 2,000
trials, corresponding to an overall success rate of 97.6\%.

\begin{table}[t]
\caption{Per-task success rates (\%) of TurboVLA on the four LIBERO suites.}
\label{tab:libero_per_task}
\begin{center}
\footnotesize
\setlength{\tabcolsep}{3.5pt}
\renewcommand{\arraystretch}{1.08}
\begin{tabular}{@{}l|cccccccccc|c@{}}
\toprule
\multicolumn{1}{c|}{\textbf{Object}}
& \textbf{0} & \textbf{1} & \textbf{2} & \textbf{3} & \textbf{4}
& \textbf{5} & \textbf{6} & \textbf{7} & \textbf{8} & \textbf{9}
& \textbf{Avg.} \\
\midrule
TurboVLA
& 100.0 & 100.0 & 100.0 & 100.0 & 100.0
& 100.0 & 100.0 & 100.0 & 100.0 & 100.0
& 100.0 \\
\midrule
\multicolumn{1}{c|}{\textbf{Goal}}
& \textbf{0} & \textbf{1} & \textbf{2} & \textbf{3} & \textbf{4}
& \textbf{5} & \textbf{6} & \textbf{7} & \textbf{8} & \textbf{9}
& \textbf{Avg.} \\
\midrule
TurboVLA
& 100.0 & 100.0 & 100.0 & 94.0 & 100.0
& 96.0 & 98.0 & 100.0 & 100.0 & 94.0
& 98.2 \\
\midrule
\multicolumn{1}{c|}{\textbf{Spatial}}
& \textbf{0} & \textbf{1} & \textbf{2} & \textbf{3} & \textbf{4}
& \textbf{5} & \textbf{6} & \textbf{7} & \textbf{8} & \textbf{9}
& \textbf{Avg.} \\
\midrule
TurboVLA
& 100.0 & 98.0 & 100.0 & 96.0 & 90.0
& 96.0 & 94.0 & 100.0 & 98.0 & 98.0
& 97.0 \\
\midrule
\multicolumn{1}{c|}{\textbf{Long}}
& \textbf{0} & \textbf{1} & \textbf{2} & \textbf{3} & \textbf{4}
& \textbf{5} & \textbf{6} & \textbf{7} & \textbf{8} & \textbf{9}
& \textbf{Avg.} \\
\midrule
TurboVLA
& 86.0 & 100.0 & 94.0 & 98.0 & 98.0
& 92.0 & 92.0 & 100.0 & 94.0 & 98.0
& 95.2 \\
\bottomrule
\end{tabular}
\end{center}
\end{table}

\subsection{Qualitative Rollouts}
\label{app:qualitative_rollouts}

\begin{figure}[t]
    \centering
    \includegraphics[width=\linewidth]{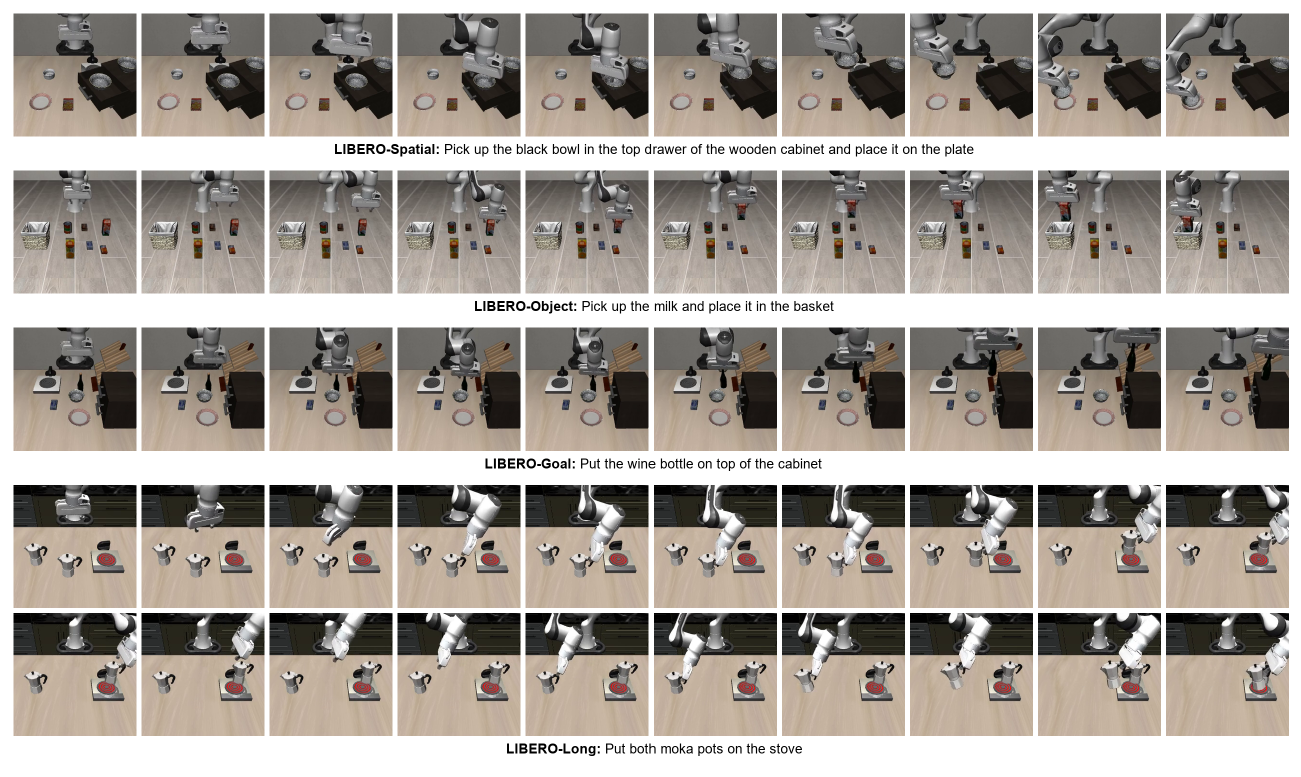}
    \caption{Qualitative rollouts on LIBERO. From top to bottom, the
    sequences show tasks from the Spatial, Object, Goal, and Long
    suites. Task instructions appear below their corresponding sequences.}
    \label{fig:libero_qualitative}
\end{figure}

\begin{figure}[t]
    \centering
    \includegraphics[width=\linewidth]{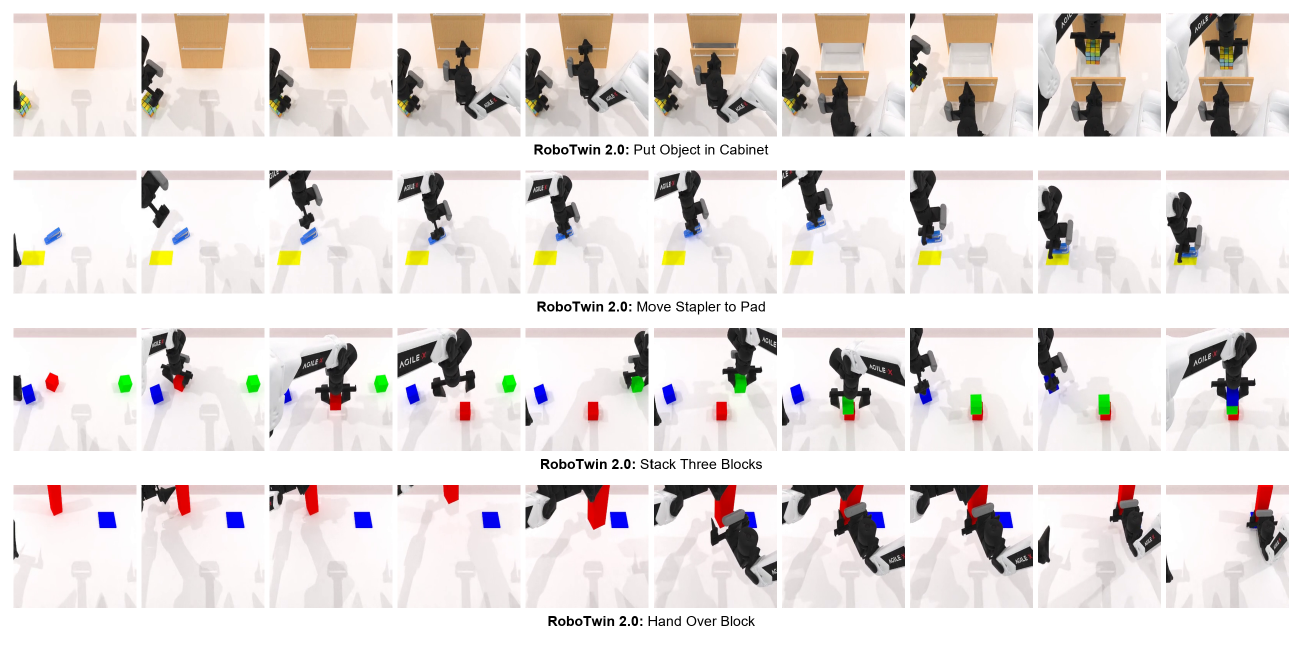}
    \caption{Qualitative rollouts on RoboTwin 2.0. From top to bottom,
    the sequences show putting an object in a cabinet, moving a stapler
    to a pad, stacking three blocks, and handing over a block.}
    \label{fig:robotwin_qualitative}
\end{figure}

\begin{figure}[p]
    \centering
    \includegraphics[width=\linewidth]{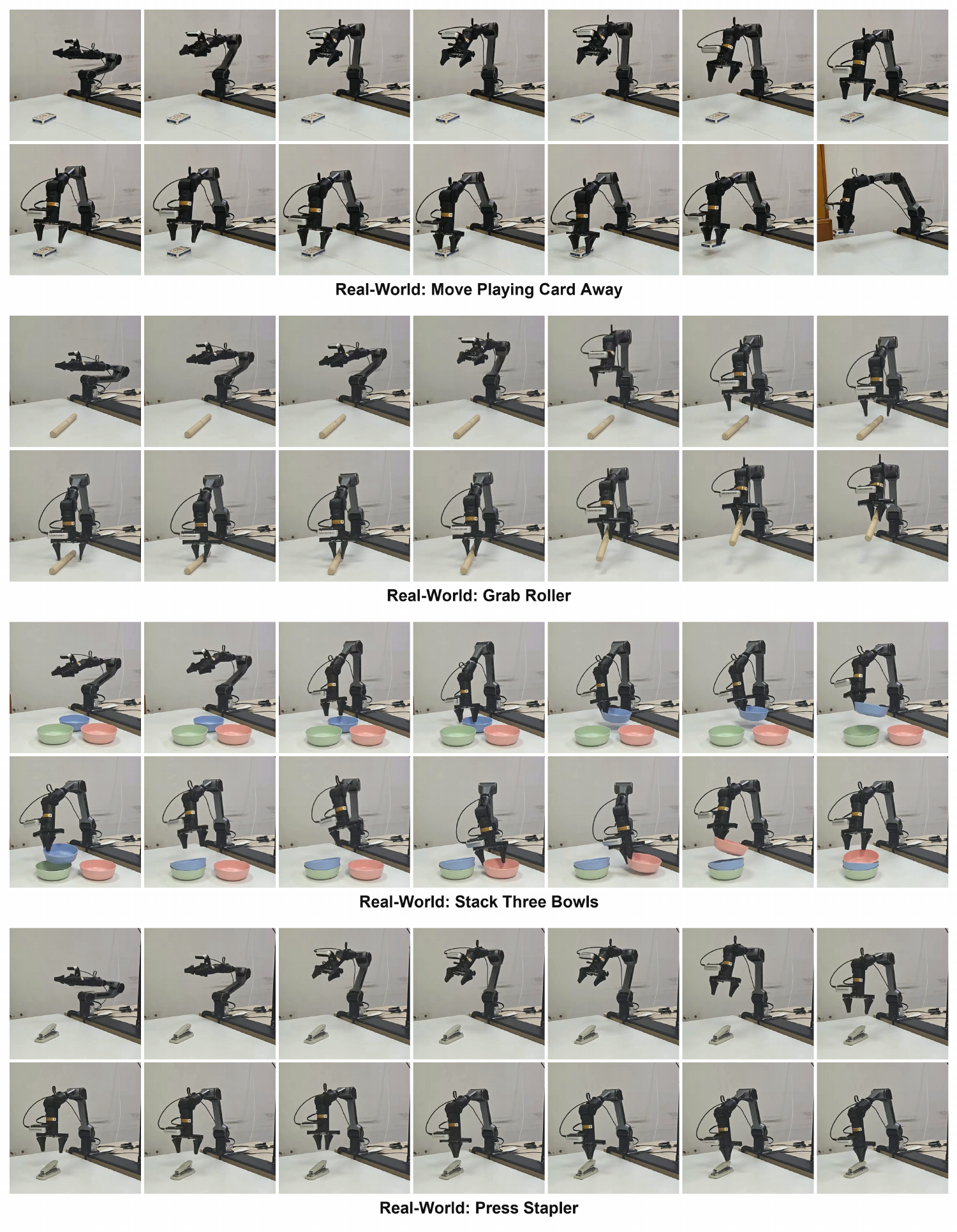}
    \caption{Real-world rollouts on the AgileX Piper platform. From top
    to bottom, the sequences show moving a playing card away, grasping
    a roller, stacking three bowls, and pressing a stapler.}
    \label{fig:realworld_qualitative}
\end{figure}

Figures~\ref{fig:libero_qualitative}--\ref{fig:realworld_qualitative}
show selected rollouts of \ourmethod in simulation and on the real
robot. The LIBERO examples span all four suites. The RoboTwin 2.0
examples cover cabinet manipulation, stapler placement, three-block
stacking, and a two-arm handover. The real-world sequences provide a
closer view of the four tasks summarized in Fig.~\ref{fig:realworld}.

\end{document}

%% file: math_commands.tex
\usepackage{amsmath,amsfonts,bm}

\def\eqref#1{equation~\ref{#1}}

\def\1{\bm{1}}

\DeclareMathAlphabet{\mathsfit}{\encodingdefault}{\sfdefault}{m}{sl}
\SetMathAlphabet{\mathsfit}{bold}{\encodingdefault}{\sfdefault}{bx}{n}

